\documentclass[11pt, oneside]{article}   	
\pdfoutput=1
\usepackage{geometry}                		
\usepackage{graphicx}				
\usepackage{amssymb}
\usepackage{amsmath}
\usepackage{inputenc}
\usepackage{bbm}
\usepackage{float}
\usepackage{authblk}
\usepackage{myharv}

\begin{document}
\bibliographystyle{kluwer}
\bibdata{../BibMaster}
\title{Parameter Learning and Change Detection Using a Particle Filter With Accelerated Adaptation}
\date{\today}
\author[*]{Karol Gellert}
\author[*]{Erik Schl\"ogl}
\affil[*]{University of Technology Sydney, Quantitative Finance Research Centre}
\renewcommand\Authands{ and }
\maketitle
\begin{abstract}
This paper presents the construction of a particle filter, which incorporates elements inspired by genetic algorithms, in order to achieve accelerated adaptation of the estimated posterior distribution to changes in model parameters. Specifically, the filter is designed for the situation where the subsequent data in online sequential filtering does not match the model posterior filtered based on data up to a current point in time. The examples considered encompass parameter regime shifts and stochastic volatility. The filter adapts to regime shifts extremely rapidly and delivers a clear heuristic for distinguishing between regime shifts and stochastic volatility, even though the model dynamics assumed by the filter exhibit neither of those features.
\end{abstract}
\newpage

\section{Introduction}
In an ideal world, using a well-specified model entails estimating the model parameters from historical data, and then applying the model with these parameters going forward, i.e. out of sample. Indeed, the bulk of the empirical academic literature in finance takes this approach.
However, practitioners' use of models, in particular for the pricing and risk management of derivative financial products relative to observed prices for liquidly traded market instruments, typically tends to depart from this ideal. Primacy is accorded to model ``calibration'' over empirical consistency, i.e., choosing a set of liquidly traded market instruments (which may include liquidly traded derivatives) as ``calibration instruments'', model parameters are determined so as to match model prices of these instruments as closely as possible to observed market prices at a given point in time. Once these market prices have changed, the model parameters (which were assumed to be constant, or at most time--varying in a known deterministic fashion) are recalibrated, thereby contradicting the model assumptions.
``Legalising'' these parameter changes by expanding the state space (e.g. via regime--switching or stochastic volatility models) shifts, rather than resolves, the problem: for example in the case of stochastic volatility, volatility becomes a state variable rather than a model parameter, and can evolve stochastically, but the parameters of the stochastic volatility process itself are assumed to be time--invariant. The limits of increasing model complexity are determined by a combination of mathematical tractability and the practicality of models.
The result is a certain disparity between empirical research and how models are used in practice. In this light we propose a practically motivated methodology in the form of an adaptive particle filter which is able to rapidly detect discrepancies between the assumed model and data including parameter changes and model mis-specification.
\paragraph{}
Particle filtering is a sequential Monte Carlo method which has become popular for its flexibility, wide applicability and ease of implementation. The origin of the particle filter is widely attributed to \citeasnoun{Gordon1993} and theirs has remained the most general filtering approach. It is an \emph{online} filtering technique ideally suited for analysing streaming financial data in a live setting. It seeks to approximate the posterior distribution of latent (unobserved) dynamic states and/or model parameters by sets of discrete sample values, where these sampled values are called ``particles.'' For a more comprehensive introduction to particle filtering see: \citeasnoun{Chen2003} for a general introduction and historical perspective; \citeasnoun{Johannes2009P}, \citeasnoun{Lopes2011}, and \citeasnoun{Creal2012} for reviews related to finance; \citeasnoun{Andrieu2004}, \citeasnoun{Cappe2007}, \citeasnoun{Chopin2010}, \citeasnoun{Kantas2009}, and \citeasnoun{Kantas2015} for parameter estimation techniques. The theoretical perspective is outlined by \citeasnoun{DelMoral2014} and covered in depth in \multicite{DelMoral2004}{DelMoral2004,DelMoral2013}.

\paragraph{}
Basic particle filter algorithms suffer from particle impoverishment, which can be broadly described as the increase in the number of zero weighted particles as the number of observations increases, resulting in fewer particles available for the estimation of the posterior.
A key distinguishing feature of most contemporary particle filters is the approach taken to deal with the problem of particle impoverishment and it continues to be a focus of effort from researchers. A variety of techniques have been proposed as a solution, the main approaches are: the use of sufficient statistics as per \citeasnoun{Storvik2002}, \citeasnoun{Johannes2007}, \citeasnoun{Polson2008}, and \citeasnoun{Carvalho2010}; maximising likelihood functions as per \citeasnoun{Andrieu2005} and \citeasnoun{Yang2008}; and random perturbation or kernel methods as per \multicite{West1993}{West1993,West1993b}, \citeasnoun{Liu2001}, \citeasnoun{Carvalho2007}, \citeasnoun{Shephard2009}, and \citeasnoun{Smith2012}.

\paragraph{}
The idea behind random perturbation, initially proposed by \citeasnoun{Gordon1993} in the context of the estimation of dynamic latent states, is that by introducing an artificial dynamic to the static parameters, the point estimates become slightly dispersed, effectively smoothing the posterior distribution and reducing the degeneracy problem. This comes at a cost of losing accuracy as the artificial dynamic embeds itself into the estimation. Motivated in part by this issue, \citeasnoun{Liu2001} introduce a random kernel with shrinking variance, a mechanism which allows for a smoothed interim posterior, but where the dispersion reduces in tandem with convergence of the posterior distribution. The method proposed by \citeasnoun{Shephard2009} is another example of this approach, introducing a perturbation to the SIR filter just prior to the resampling stage such that new samples are drawn from an already smoothed distribution, avoiding damage to the asymptotic properties of the algorithm. A common theme of these approaches is the assumption that the parameters are fixed over the observation period, whereas we will operate under the pragmatic (and somewhat self--contradictory) premise of the frequent recalibration of financial market models conducted by practitioners: Model parameters are fixed, until they change.\footnote{If desired, one could interpret this as estimation of changes in model regime, but where one remains agnostic about the number and nature of the model regimes. Particle filters have been applied to estimate regime--switching models, see for example \citeasnoun{Carvalho2007} and \citeasnoun{Bao2012}.} To this end, we adapt the idea of random perturbation to a more general parameter detection filter, pursuing a similar objective as \citeasnoun{Nemeth2014},\footnote{See also \citeasnoun{Nemeth2012}.} who develop a particle filter for the estimation of dynamically changing parameters. This type of problem is usually motivated by tracking maneuvering targets, perhaps an apt metaphor for a financial market model requiring repeated recalibration of model parameters as time moves on.

\paragraph{}
Our contribution to the literature is the introduction of a particle filter with accelerated adaptation designed for the situation where the subsequent data in online sequential filtering does not match the model posterior filtered based on data up to a current point in time. This covers cases of model misspecification as well as sudden regime changes or rapidly changing parameters. The proposed filter is an extension of on-line methods for parameter estimation which achieve smoothing using random perturbation. The accelerated adaptation is achieved by introducing a dynamic to the random perturbation parameter, allowing particle--specific perturbation variance; this combines with re-selection to produce a genetic algorithm which allows for rapid adaptation to mismatching or changing dynamics in the data. The similarity between particle filtering and genetic algorithms has been noted before, see for example \citeasnoun{Smith2012}, who use genetic algorithm mutation as a resampling step in the SIR filter. We reinforce the genetic algorithm aspect of the particle filter to detect and rapidly adapt to any discrepancies between the model and realised dynamic by exploiting random perturbations, in a sense taking the opposite direction of methods in the literature which seek to control random perturbation in order to remove biases in the estimation of parameters assumed to be \emph{fixed}.

\paragraph{}
This approach leads us to a useful indicator of when changes in model parameters are being signalled by the data. The effectiveness of this heuristic measure is based on the notion that in the case of perfect model specification no additional parameter ``learning'' would be required. We show how this indicator can provide useful information for characterising the empirical underlying dynamics without using highly complex models (meaning models which assume stochastic state variables where the simpler model uses model parameters). This allows for the use of a more basic model implementation to gain insight into more complex models, to make data--driven choices on how the simpler models might most fruitfully be extended. For example, our indicator will behave quite distinctly for an unaccounted regime change in the dynamic as opposed to an unaccounted stochastic volatility dynamic.

The remainder of the paper is organised as follows. Section 2 recalls the basic particle filter construction. Section 3 iteratively presents the evolution of the particle filter methodology based on the existing literature, culminating in the \citeasnoun{Liu2001} filter, which forms the starting point of our filter with accelerated adaptation. Section 4 presents our particle filter incorporating additional elements inspired by genetic algorithms, adding this elements step by step and providing examples demonstrating their effectiveness. Section 5 concludes.

\section{Particle filter}
\subsection{General framework}
Consider a sequence of Markovian discrete time states $x_{1:t}=\lbrace{}x_1,...,x_t\rbrace$ and discrete observations $y_{1:t}=\lbrace{}y_1,...,y_t\rbrace$. In general, particle filtering is concerned with the filtering problem characterised by state-space equations:\footnote{using notation from \citeasnoun{Chen2003}}
\begin{gather}
x_{t+1}=f(x_t,d_t)\label{state_transition}\\
y_t={g}(x_t,v_t)\label{measurement_equation}
\end{gather}
with $d_t$ and $v_t$ as independent random sequences in the discrete time domain. The \emph{transition density} of the state $p(x_{t+1}|x_t)$ and the observation \emph{likelihood} $p(y_t|x_t)$ are obtained from equations \eqref{state_transition} and \eqref{measurement_equation}. The objective of particle filtering is the sequential estimation of the posterior density $p(x_t|y_{1:t})$ which can be expressed in the recursive Bayesian setting:\footnote{see \citeasnoun{Chen2003} for derivation}
\begin{gather}
p(x_t|y_{1:t})=\frac{p(y_t|x_t)p(x_t|y_{1:t-1})}{p(y_t|y_{1:t-1})}
\end{gather}
where $p(x_t|y_{1:t-1})$ is the \emph{prior} and is determined by the integral:
\begin{gather}
p(x_t|y_{1:t-1})=\int{}p(x_t|x_{t-1})p(x_{t-1}|y_{1:t-1})dx_{t-1}\label{prior_equation}
\end{gather}
$p(y_t|y_{1:t-1})$ is the \emph{evidence} and is determined by:
\begin{gather}
p(y_t|y_{1:t-1})=\int{}p(y_t|x_t)p(x_t|y_{1:t-1})dx_t
\end{gather}

The continuous posterior is approximated in the particle filter by a discrete random measure $\lbrace{}x_t^{(i)},\pi_t^{(i)}\rbrace{}^N_{i=1}$ composed of sample values of the state $x_t^{(i)}$ with associated weights $\pi_t^{(i)}$, and $N$ denoting the total number of particles\footnote{see \citeasnoun{Li2015}} as follows:
\begin{gather}\label{posteriorEQ}
p(x_t|y_{1:t})\approx\sum\limits_{i=1}^N\delta_{\{x_t^{(i)}=x_t\}}\pi_t^{(i)}
\end{gather}
where $\delta$ is the Dirac delta function.

\subsection{Filtering for parameters with directly observed states}
\paragraph{}
The general state-space framework is a combination of state and observation dynamics. The observation dynamics are useful in problems which require modelling of observation uncertainty, for example applications involving physical sensors. In empirical finance the observation usually consists of a directly observed price, which often is assumed not have any inherent observation uncertainty. It does not mean state uncertainty cannot be modelled but here unobserved states are artefacts of the modelling assumptions rather than physical quantities of the system --- stochastic volatility is one such example.
\paragraph{}
In this paper we focus on estimation of only model parameters, in order to present our approach in the simplest setting where states are assumed to be directly observed --- by adapting the results in existing literature in the appropriate manner, our approach would be straightforward to extend to the estimation of latent state variables. We begin with a derivation of the recursive Bayesian framework below. Similarly to \citeasnoun{Chen2003}, the last step of the derivation relies on the assumption that the states follow a first-order Markov process and therefore $p(x_t|x_{1:t-1}) = p(x_t|x_{t-1})$. Let $\theta$ represent the parameter set, the parameter posterior is
\begin{align}
p(\theta|x_{1:t+1})&=\frac{p(x_{1:t+1}|\theta)p(\theta)}{p(x_{1:t+1})}\\
&=\frac{p(x_{t+1},x_{1:t}|\theta)p(\theta)}{p(x_{t+1},x_{1:t})}\\
&=\frac{p(x_{t+1}|x_{1:t},\theta)p(x_{1:t}|\theta)p(\theta)}{p(x_{t+1}|x_{1:t})p(x_{1:t})}\\
&=\frac{p(x_{t+1}|x_{1:t},\theta)p(\theta|x_{1:t})}{p(x_{t+1}|x_{1:t})}\\
&=\frac{p(x_{t+1}|x_{1:t},\theta)p(\theta|x_{1:t})}{\int{}p(x_{t+1}|x_{1:t},\theta)p(\theta|x_{1:t})d\theta}\\
&=\frac{p(x_{t+1}|x_t,\theta)p(\theta|x_{1:t})}{\int{}p(x_{t+1}|x_t,\theta)p(\theta|x_{1:t})d\theta}
\end{align}
The above formulation establishes a recursive relationship between sequential posteriors. The particle approximation is based on the discretisation of $\theta^{(i)}$ as shown below. Introducing weight notation $\pi_{t+1}^{(i)}:=p(\theta^{(i)}|x_{1:t+1})$ and the un-normalised weight $\hat{\pi}_{t+1}^{(i)}$ we have

\begin{gather}\label{general_estimate_equation}
\pi_{t+1}^{(i)}\approx{}\frac{p(\bold{x}_{t+1}|\bold{x}_t,\theta^{(i)})\pi_t^{(i)}}{\sum\limits_{i=1}^{N}p(\bold{x}_{t+1}|\bold{x}_t,\theta^{(i)})\pi_t^{(i)}}=\frac{\hat{\pi}_{t+1}^{(i)}}{\sum\limits_{i=1}^{N}\hat{\pi}_{t+1}^{(i)}}
\end{gather}

After initialisation a basic filtering algorithm consists of iterative application of two steps to calculate the above approximation. The \emph{update} step is the calculation of $\hat{\pi}_{t+1}^{(i)}$ for all $i$, and the \emph{normalisation} step obtains the posterior estimates $\pi_{t+1}^{(i)}$.

\section{Implementation and numerical results}
From this point, the paper follows an iterative approach to demonstrate the evolution of particle filter methodology as it exists in the current literature. Each iteration consists of a definition of a particle filter algorithm, followed by a simulation study  with a focus on deficiencies which are used to motivate the next innovation. Each incremental addition to the filter algorithm aims to resolve the deficiency found in the previous step. Thus this section focuses on already existing techniques, leading up to the next section which contains the main contributions of this paper.
This incremental process is initialised with the most basic algorithm and simulation model chosen to represent a minimal implementation of a particle filter.
The choice of a simple simulation model makes available a known posterior distribution, providing a benchmark for measuring the performance of the filter. To measure performance we use the Kolmogorov--Smirnov statistic. Although this is not a common choice, we find it particularly useful for measuring performance against the benchmark, as well providing an intuitive measure for demonstrating particle impoverishment.
\subsection{Preliminaries}
\subsubsection{Observation process}
Begin with a basic Gaussian stochastic process, defined by the stochastic differential equation
\begin{gather}\label{gaussian_sde}
dx_t=\sigma{}dW_t
\end{gather}
where $W_t$ denotes a standard Wiener process. Discrete observations used for the simulation study are generated using the Euler-Maruyama discretisation, i.e.
\begin{gather}
\Delta{}x_t=x_t-x_{t-1}=\sigma\Delta{}W_t
\end{gather}
The particle filter presented in this section will be concerned with estimating the posterior of $p(\sigma_t|x_{1:t})$ given a set of observations $x_{1:t}$ generated by the above process. Note the use of subscript in $\sigma_t$ to associate the estimate with data up to time $t$.\\
The transition density for this process is given by:
\begin{gather}\label{GaussTransitionDensity}
p(x_t|x_{t-1},\sigma)=\frac{1}{\sqrt{2\pi\sigma^2}}e^\frac{-{\Delta{}x_t}^2}{2\sigma_t^2}
\end{gather}

\subsubsection{Benchmark posterior}
Finding the posterior of $\sigma$ from observations generated by the above process is equivalent to finding the posterior distribution of the variance given a set of Gaussian increments $\Delta{}x_t$. An established result, based on Cochran's theorem\footnote{see \citeasnoun{Cochran1934}}, states that the distribution of $\sigma^2_t$ is obtained from the chi-square distribution with $n-1$ degrees of freedom; $\chi_{n-1}^2$ according to:
\begin{gather}
\frac{\hat\sigma^2_tn}{\sigma^2_t}\sim{}\chi_{n-1}^2
\end{gather}
where $\hat\sigma^2_t$ is the maximum likelihood estimator; $\hat\sigma^2_t=\frac{\sum_t{\Delta{}x_t^2}}{n}$ with $n$ the number of observations.
The theoretical posterior can be written as:
\begin{gather}\label{chidensity}
p(\sigma_t|x_{1:t})=\chi{}^2_{n-1}\left(\frac{\hat\sigma^2_tn}{\sigma^2_t}\right)
\end{gather}
with cumulative distribution function (CDF):
 \begin{gather}
 F(\sigma_t)=\int\limits_0^{\sigma_t}\chi{}^2_{n-1}\left(\frac{\hat\sigma^2_tn}{\sigma^2_t}\right)d\sigma_t
 \end{gather}

\subsubsection{Measuring performance}
Assessing the performance of the particle filter in the presence of a theoretical benchmark amounts to measuring the distance between two posterior distributions with respect to increasing number of particles and increasing number of observations. The Kolmogorov-Smirnov (KS) statistic, which measures the maximum distance between CDFs, seems a natural choice, however it is not commonly used in literature, with \citeasnoun{Djuric2010} providing one of the few examples of usage related to the particle filter.
Define the estimated CDF as $$F^*(\sigma_t)=\sum\limits_i{}\mathbbm{I}_{(\sigma_t^{(i)}\leq\;\sigma_t)}\pi_t^{(i)}.$$
The KS statistic measures the maximum distance between the theoretical $F(\sigma_t)$ and estimated posterior $F^*(\sigma_t)$:
 \begin{gather}\label{ks_statistic}
KS=\sup_{\sigma_t}|F^*(\sigma_t)-F(\sigma_t)|
 \end{gather}
It must be stressed that this approach is limited in use for this specific case due to the availability of a known posterior. Within this limitation, it is a simple and effective approach for demonstrating convergence, as well as demonstrating the issue of particle impoverishment; including the efficacy of the Liu and West filter for resolving it.
\subsubsection{Convergence}
Convergence has been the subject of significant research and a wide range of results exist in the literature. For detailed theoretical analysis see \citeasnoun{Chopin2004},  \citeasnoun{DelMoral2014}, \citeasnoun{Douc2007} and \citeasnoun{Doucet2008}. For a comprehensive survey of convergence results refer to \citeasnoun{Crisan2002}.
The present paper employs numerical testing of expected convergence results as a means to assess particle filter performance. The numerical tests are based on the KS statistic relying on the basic assertion that convergence of the KS statistic implies convergence of the estimated posterior to the benchmark distribution:
\begin{gather}
\underset{\sigma_t}{\sup}|F^*(\sigma_t)-F(\sigma_t)|\xrightarrow[N\rightarrow{}\infty{}]{}0\implies{}F^*(\sigma_t)\xrightarrow[N\rightarrow{}\infty{}]{}F(\sigma_t)
\end{gather}
Additionally, the KS statistic is used to assess convergence with respect to the number of observations, where it proves to be an effective indicator of a known issue with particle filters referred to as particle impoverishment\footnote{see \citeasnoun{Chen2003} p. 26 for an introduction to particle impoverishment}.
\subsection{Basic Filter (SIS)}
\begin{figure}[t]
\centerline{\includegraphics[scale=0.75]{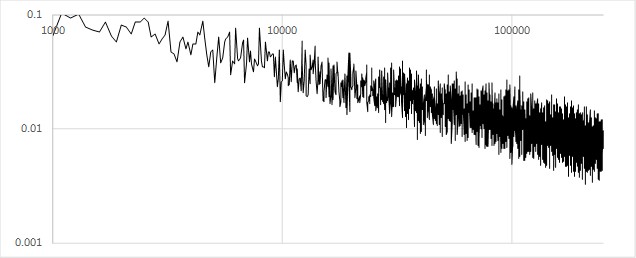}}
\caption{KS statistic for increasing number of particles (logarithmic scales)}\label{convergence_random}
\end{figure}

\begin{figure}[t]
\centerline{\includegraphics[scale=0.75]{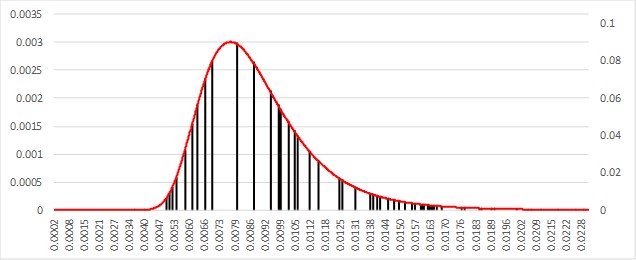}}
\caption{theoretical (red, lhs) and estimated (black, rhs) $\sigma$ (horizontal) posterior.}\label{pdf_random}
\end{figure}

\begin{figure}[t]
\centerline{\includegraphics[scale=0.75]{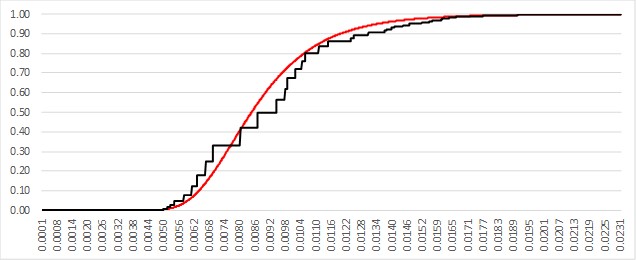}}
\caption{theoretical (red) and estimated (black) $\sigma$ (horizontal) posterior CDF.}\label{cdf_random}
\end{figure}

A basic filtering algorithm consists of initialisation followed by iterative application of two steps corresponding to \eqref{general_estimate_equation}.
It is essentially an adaptation of the filter known as sequential importance sampling (SIS) for the detection of a static parameter. A filter for dynamic state variables also would include a draw from the state variable transition kernel, see \citeasnoun{Chen2003}.
The \emph{update} step is the calculation of $\hat{\pi}_{t+1}^{(i)}$ for all $i$, and the \emph{normalisation} step obtains the posterior estimates $\pi_{t+1}^{(i)}$. The algorithm is defined as follows: \\[1ex]
\noindent
\fbox{\begin{minipage}{38.5em}
1: \textit{Initilisation} For each particle; draw $N$ particles $\sigma_0^{(i)}\sim{}U(a,b)$ and $\pi_0^{(i)}=\frac{1}{N}$\\
2: Sequentially for each observation:\\
\indent{}	2.1: \textit{Update} For each particle update weight $\hat{\pi}_t^{(i)}=\pi_t^{(i)}p(x_t|x_{t-1},\sigma_t^{(i)})$\\
\indent{}   2.2: \textit{Normalisation} For each particle $\pi_t^{(i)}=\frac{\hat{\pi}_t^{(i)}}{\sum{}\hat{\pi}_t^{(i)}}$
\end{minipage}}\\[1ex]
The performance of the filter is tested numerically by rerunning it with an increasing number of particles given a fixed set of observations, and recording the KS statistic for each run. The results are presented in Figure \ref{convergence_random}, showing the value of the KS statistic with respect to the number of particles, both in log-space. Convergence is evident, however the most prominent aspect of the results is instability as $N\rightarrow{}\infty$. The instability indicates that while there is an overall convergence trend, incremental increases in the number of particles result in significant noise in the KS statistic.
To better understand the reason behind the instability, the theoretical \eqref{chidensity} and the estimated posterior \eqref{posteriorEQ} PDF and CDF are compared visually for a small number of particles. The PDF comparison is shown in Figure \ref{pdf_random}.
Each vertical line in the plot represents a particle weight $\pi_t^{(i)}$ and is shown against the theoretical posterior\footnote{Note the estimate and theoretical posterior are shown at different scales as the scaling of the estimate depends on the number of particles, only converging to the scaling of the theoretical posterior as $N\rightarrow{}\infty$}. The scaled comparison demonstrates a very good correspondence between the estimated and theoretical shape of the PDF. It is also evident that the estimation points are unevenly distributed, reflecting the randomly initialised particle locations.
The estimated and theoretical CDFs are compared in Figure \ref{cdf_random}. The KS statistic corresponds to the maximum vertical distance between the two plots. Intuitively it appears that the vertical distance between the CDFs is related to the horizontal distance between adjacent estimation points, identifiable in the CDF as the end points of each piecewise flat interval. It follows that the maximum vertical distance, i.e the KS statistic, is related to the maximum distance between adjacent estimation points. The random initialisation of the estimation points means that the maximum distance between adjacent points does not decrease monotonically as $N\rightarrow{}\infty$, resulting in the instability observed in the KS statistic.

\subsection{Equal Spacing}
\begin{figure}[t]
\centerline{\includegraphics[scale=0.75]{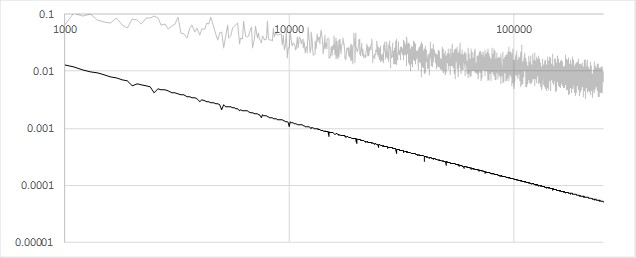}}
\caption{KS statistic or increasing number of particles (logarithmic scales), random (grey) vs equally spaced (black) distribution of estimation point densities}\label{convergence_random_uniform}
\end{figure}

\begin{figure}[t]
\centerline{\includegraphics[scale=0.75]{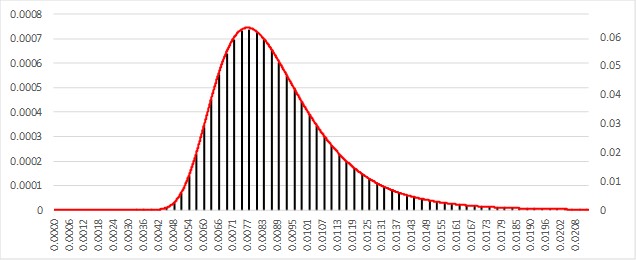}}
\caption{theoretical (red, lhs) and estimated (black, rhs) $\sigma$ (horizontal) posterior for equally spaced estimation points}\label{pdf_uniform}
\centerline{\includegraphics[scale=0.75]{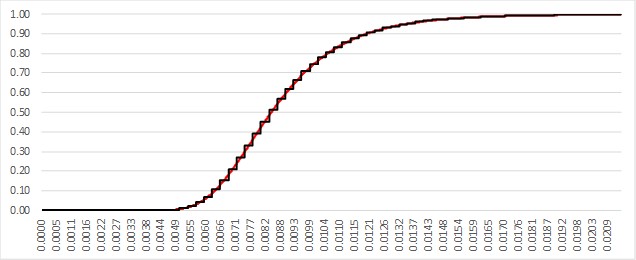}}
\caption{theoretical (red) and estimated (black) $\sigma$ (horizontal) posterior CDF for equally spaced estimation points}\label{cdf_uniform}
\end{figure}

\begin{figure}[t]
\centerline{\includegraphics[scale=0.75]{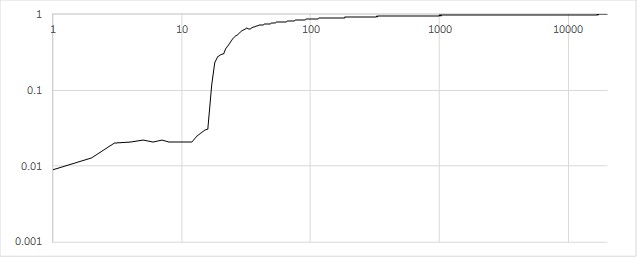}}
\caption{Proportion of zero weight particles with increasing number of observations for the SIS filter in log-space}\label{zero_weight_particles}
\end{figure}

The above reasoning leads to a trivial improvement: if the point density locations are equally spaced the maximum distance between adjacent points will decrease monotonically as $N\rightarrow{}\infty$. Further, even spacing guarantees that this maximum distance is minimised for any given number of particles\footnote{This does not guarantee the optimal estimation point distribution with respect to the KS statistic, but the improvement in performance is substantial}.The initialisation in the filtering algorithm is altered to reflect equally spaced estimation points:\\[1ex]
\noindent
\fbox{\begin{minipage}{38.5em}
1: \textit{Initilisation} For each particle; let $\sigma_0^{(i)}=\frac{(b-a)i}{N}$ and $\pi_0^{(i)}=\frac{1}{N}$\\
2: Sequentially for each observation:\\
\indent{}	2.1: \textit{Update} For each particle update weight $\hat{\pi}_t^{(i)}=\pi_t^{(i)}p(x_t|x_{t-1},\sigma_t^{(i)})$\\
\indent{}   2.2: \textit{Normalisation} For each particle $\pi_t^{(i)}=\frac{\hat{\pi}_t^{(i)}}{\sum{}\hat{\pi}_t^{(i)}}$
\end{minipage}}\\[1ex]
The convergence test from the previous section is rerun with the above adjustment. The result, shown in Figure \ref{convergence_random_uniform}, clearly demonstrates that this simple change has resulted in a faster and more stable rate of convergence.\footnote{
This is somewhat related to research focusing on sequential monte carlo using quasi-random rather than pseudo-random draws, see for example \citeasnoun{Gerber2015}}
\footnote{notably the convergence is linear in log-space suggesting the form $\sup|F^*(\sigma)-F(\sigma)|\leq{}\frac{C}{N}$ echoing theoretical results for the convergence of mean square error, see \citeasnoun{Crisan2002}, section V}
The reason for the improvement is evident in the PDF Figure \ref{pdf_uniform} and CDF Figure \ref{cdf_uniform} and confirms the assertion from the previous section. The shape of the PDF is preserved at equally spaced intervals which allows closer and more consistent alignment between the CDFs, resulting in the elimination of noise from the convergence of the KS statistic.
\paragraph{}
A well known problem with the basic particle filter is that the number of particles with non-zero weights can only decrease with each iteration\footnote{see \citeasnoun{Chen2003} p. 26 for a good summary}. Zero weights occur when the estimated posterior probability at a particular estimation point falls below the smallest positive floating point number available for the computing machine on which the filter is implemented. To demonstrate the problem, the proportion of zero weighted particle is plotted against the number of observations in Figure \ref{zero_weight_particles}.
 This \emph{weight degeneration} is a problem particularly for detection of dynamic state variables where it essentially diminishes the sample domain. In this case the problem is largely solved by introducing a resampling step where the zero weight particles are replaced by sampling from the non-zero weight particles according to their relative estimated probabilities. For detection of static model parameters, simply resampling only removes the zero weight particles but does not resolve the sample impoverishment problem. Because the model parameters are assumed to be fixed, resampling simply concentrates more particles on the same (diminishing) number of estimation points. The resolution of this issue for static parameters will be demonstrated over the next two sections.
\subsection{Resampling: SIR Filter}
\begin{figure}[t]
\centerline{\includegraphics[scale=0.8]{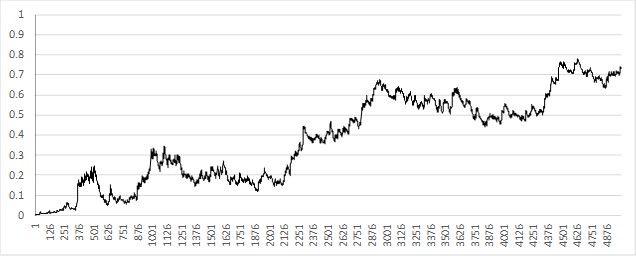}}
\caption{KS statistic for increasing number of observations for the SIR filter}\label{impoverishment_sir}
\end{figure}

\begin{figure}[t]
\centerline{\includegraphics[scale=0.8]{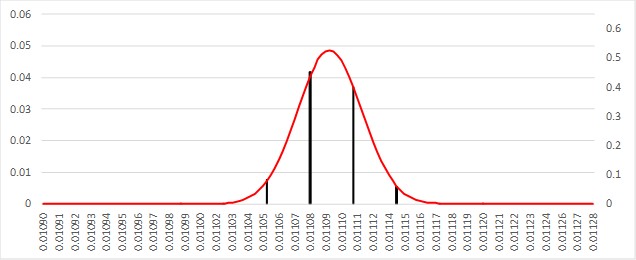}}
\caption{theoretical (red, lhs) and estimated (black, rhs) $\sigma$ (horizontal) posterior demonstrating particle impoverishment.}\label{pdf_impoverished_sis}
\end{figure}

The first step to reducing particle impoverishment is to redistribute the particles according to the current posterior estimate, a technique called sample importance resampling (SIR). This redistribution of particles replicates higher probability particles and discards any with zero or very low weighting, thus resolving the problem highlighted in the previous section. Various methods for resampling have been proposed in literature. This paper uses systematic resampling, which can be found in the survey analysis of  \citeasnoun{Hol2006}, where it is described as having the lowest discrepancy and reduced computation complexity without deterioration of the estimate.\footnote{Our own experimentation confirms these findings.}
The goal of the resampling step is to transform the posterior distribution approximated by $N$ particles of differing weights into one approximated by $N$ particles of equal weight.
The resampling begins by generating $N$ ordered numbers:
\begin{gather}
u_k=\frac{(k-1)+\tilde{u}}{N}\;\text{, with}\;\tilde{u}\sim{}U[0,1),
\end{gather}
then reselecting the particles according to
\begin{gather}
\sigma_t^{(k)}=\sigma_t^{(i)}\;\text{, with}\;i\;\text{s.t.}\;u_k\in\Bigg[\sum\limits_{s=1}^{i-1}\pi_t^{(s)},\sum\limits_{s=1}^i\pi_t^{(s)}\Bigg)
\end{gather}
The filtering algorithm is adjusted to include the resampling step:\\[1ex]
\noindent
\fbox{\begin{minipage}{38.5em}
1: \textit{Initialisation} For each particle; let $\sigma_0^{(i)}=\frac{(b-a)i}{N}$ and $\pi_0^{(i)}=\frac{1}{N}$\\
2: Sequentially for each observation:\\
\indent{}	2.1: \textit{Update} For each particle update weight $\hat{\pi}_t^{(i)}=\pi_t^{(i)}p(x_t|x_{t-1},\sigma_t^{(i)})$\\
\indent{}   2.2: \textit{Normalisation} For each particle $\pi_t^{(i)}=\frac{\hat{\pi}_t^{(i)}}{\sum{}\hat{\pi}_t^{(i)}}$\\
\indent{}   2.3: \textit{Resampling} Generate a new set of particles: $$p(\sigma_t|x_{1:t})\approx\sum\limits_{i=1}^N\delta_{\{\sigma_t^{(i)}=\sigma_t\}}\pi_t^{(i)}\xrightarrow[resample]{}p(\sigma_t|x_{1:t})\approx\sum\limits_{k=1}^N\frac{1}{N}\delta_{\{\sigma_t^{(k)}=\sigma_t\}}$$
\end{minipage}}\\[1ex]
As discussed, the above algorithm discards zero weighted particles (or, more accurately, is likely to discard particles of very low weight), resolving the problem posed in the previous section. However, for static model parameter estimation the particle impoverishment still persists, because resampling simply concentrates particles on the same estimation points. Because the parameters are assumed to be fixed, their value does not change as it would for dynamic state parameters. As the number of observations increases, the theoretical posterior density becomes increasingly more  concentrated around the maximum likelihood estimate, in the limit approaching the Dirac delta measure.
The key intuition to understanding particle impoverishment is that the estimated posterior will concentrate the estimation density on a single point after a finite number of observations rather than in the limit: On each update of the weight $\pi_t^{(i)}$, the total weight becomes more and more concentrated on the parameter value $\theta^{(i)}$ of maximum likelihood given the observations, but for a finite number of observations the parameter value of maximum likelihood does not necessarily coincide with the ``true'' parameter of the data generating process, and for a finite number of particles, the discretisation of the parameter space by the $\theta^{(i)}$ will also mean that the best possible $\theta^{(i)}$ will not coincide with the exact ``true'' parameter.
Therefore, in the limit the theoretical and estimated posterior will both be concentrated on a single point at different locations, with the estimate reaching this state after a finite amount of observations. Once the estimate reaches this point the KS statistic is based on the one estimation point (denote as $\sigma^*$) and equals $\max(1-F(\sigma^*),F(\sigma^*))$. As the theoretical posterior converges, its PDF narrows until the single remaining estimation point is outside its numerically significant domain; reflecting this the KS statistic approaches the maximum value 1.
This is demonstrated numerically by running the particle filter over a large number observations and recording the KS statistic at each sequential estimate, as shown in Figure \ref{impoverishment_sir}.
The KS statistic approaching 1 as the number of observations increases is an indicator of theoretical and particle filter posteriors diverging from each other as a consequence of particle impoverishment.
As an additional explanation, Figure \ref{pdf_impoverished_sis} shows an example of a PDF for an impoverished state of the particle filter, where the posterior is estimated by just four particle locations. The example shows a state where the theoretical posterior has narrowed as it converges, however the spacing between the estimation points has not changed. This will eventually lead to the posterior being estimated by a single point, which eventually will fall outside of the theoretical posterior.
One of the many approaches motivated by the problem of particle impoverishment, initially proposed by \citeasnoun{Gordon1993}, is to add small random perturbations to every particle at each iteration of the filtering algorithm, that is
\begin{gather}
\theta_{t+1}=\theta_t+\zeta_{t+1}\\
\zeta_{t+1}\sim{}N(0,W_{t+1})
\end{gather}
where $W_{t+1}$ is a specified variance matrix.
While this approach provides a framework for addressing particle impoverishment, it does so at the cost of accuracy to the posterior distribution. Any random perturbation to the fixed parameters introduces an artificial dynamic resulting in potential overdispersion of the parameter estimate.
For example, if the variance $W_{t+1}$ of the random perturbation is constant, the constant value becomes the minimum variance of the posterior estimate, i.e at some point the minimum variance becomes larger than the variance of the theoretical posterior, almost exactly the opposite effect to particle impoverishment.
It is therefore desirable to have the perturbation variance shrink in line with the posterior convergence such that it always remains only a relatively small contributor to the estimation variance.
One such approach which explicitly addresses overdispersion is proposed by \citeasnoun{Liu2001}, and the literature refers to this as the Liu and West filter.
\subsection{Liu and West filter}
\begin{figure}[t]
\centerline{\includegraphics[scale=0.8]{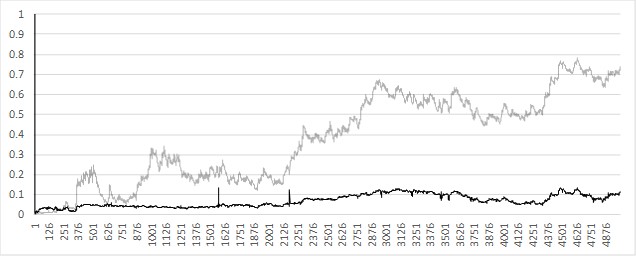}}
\caption{KS statistic for increasing number of observations SIR (grey) and Liu and West (black)}\label{impoverishment_perturb}
\end{figure}

\begin{figure}[t]
\centerline{\includegraphics[scale=0.8]{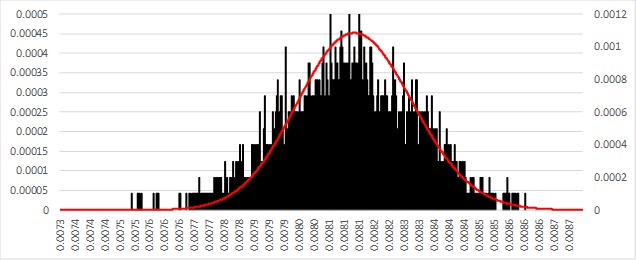}}
\caption{theoretical (red, lhs) and estimated (black, rhs) posterior for $\sigma$ (horizontal) in the Liu and West filter}\label{pdf_perturb}
\end{figure}

To resolve the problem of over-dispersion, \citeasnoun{Liu2001} put forward an approach using a kernel interpretation of the random perturbation proposed by \citeasnoun{Gordon1993}. The idea of the kernel representation is that each parameter in the particle population exists as a density instead of a single point. The overdispersion is resolved by linking the variance of the kernel to the estimated posterior variance such that it shrinks proportionally to the convergence of the estimated posterior.
The practical application within the filter algorithm is to draw the parameter from the kernel density for each particle at each iteration. The kernel is expressed as a normal density ${\mathcal{N}}(\sigma|m,S)$ with mean $m$ and variance $S$ and replaces the Dirac delta density in equation \eqref{posteriorEQ}:
\begin{gather}\label{LWKernel}
p(\sigma_t|x_{1:t})\approx\sum\limits_{i=1}^N\pi_t^{(i)}{\mathcal{N}}(\sigma_t^{(i)}|m_t^{(i)},h^2V_t)
\end{gather}
where $V_t$ is the variance of the current posterior $V_t=\frac{1}{N}\sum\limits_{i}(\sigma_t^{(i)}-\overline{\sigma}_t)^2$ and
\begin{gather}
m_t^{(i)}=c\sigma_t^{(i)}+(1-c)\overline{\sigma}_t
\end{gather}
with $c=\sqrt{1-h^2}$ and $\overline{\sigma}_t$ the mean of the current posterior.
The filtering algorithm now includes a kernel smoothing step where the posterior points are drawn from the kernel defined in eq. \eqref{LWKernel}: \\[1ex]
\noindent
\fbox{\begin{minipage}{38.5em}
1: \textit{Initialisation} For each particle; let $\sigma_0^{(i)}=\frac{(b-a)i}{N}$ and $\pi_0^{(i)}=\frac{1}{N}$\\
2: Sequentially for each observation:\\
\indent{}	2.1: \textit{Update} For each particle update weight $\hat{\pi}_t^{(i)}=\pi_t^{(i)}p(x_t|x_{t-1},\sigma_t^{(i)})$\\
\indent{}   2.2: \textit{Normalisation} For each particle $\pi_t^{(i)}=\frac{\hat{\pi}_t^{(i)}}{\sum{}\hat{\pi}_t^{(i)}}$\\
\indent{}   2.3: \textit{Resampling} Generate a new set of particles: $$p(\sigma_t|x_{1:t})\approx\sum\limits_{i=1}^N\delta_{\{\sigma_t^{(i)}=\sigma_t\}}\pi_t^{(i)}\xrightarrow[resample]{}p(\sigma_t|x_{1:t})\approx\sum\limits_{k=1}^N\frac{1}{N}\delta_{\{\sigma_t^{(k)}=\sigma_t\}}$$
\indent{}   2.4: \textit{Kernel smoothing} For each particle apply $\sigma_t^{(i)}\sim{}{\mathcal{N}}(\sigma_t^{(i)}|m_t^{(i)},h^2V_t)$
\end{minipage}}\\[1ex]
\noindent
Numerical results demonstrating the effectiveness of the Liu and West filter in reducing particle impoverishment are shown in Figure \ref{impoverishment_perturb}. The KS statistic for the Liu and West filter remains relatively constant as the number of observations increases, indicating that the filter estimate convergence with respect to the number of observations is well aligned with the theoretical posterior.
The reason for the improvement is confirmed by comparison of the estimated posterior to the theoretical PDF, shown in Figure \ref{pdf_perturb}. The example demonstrates the effectiveness of  the kernel in estimating both the location and variance of the theoretical posterior.

\section{Parameter learning and change detection}
In this section we continue to extend the particle filter, introducing techniques representing the main research contribution of this paper.
The section begins by introducing a regime shift to the Gaussian reference model in order to pose a more difficult filtering problem and highlight the adaptive aspect of the Liu and West filter.
We show that the adaptation is a result of the combination of random perturbation and re-selection, forming a genetic algorithm capable of adapting to parameter regime changes.
After demonstrating the link between adaptation speed and the size of the random perturbation, we propose an extension to the Liu and West filter which increases the kernel variance when required for adaptation. This is achieved by exploiting the genetic algorithm embedded in the Liu and West filter, thus allowing the size of parameter random perturbation to evolve as part of the already existing process. The result is a filter capable of adapting to regime changes and converging to the Liu and West filter when adaptation is not required. We also illustrate the capacity of the filter to adapt to stochastic volatility.
Finally we demonstrate how measuring the average adaptation at each iteration can provide useful information which can be used to distinguish between different dynamics of the underlying data.

\subsection{Regime shift} \label{RegimeShift}
\begin{figure}[t]
\centerline{\includegraphics[scale=0.8]{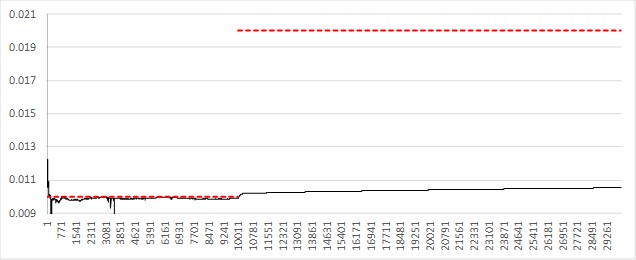}}
\caption{Estimated expected value (black) vs simulation input value (red) of $\sigma_t$ with regime change after 10,000 steps for the Liu and West filter}\label{t_convergence_regime_lw}
\end{figure}

Consider a model where at time $t=t^*$ there is a change in volatility,
\begin{gather}
dx_t=(\sigma_1\mathbbm{I}_{t<t^*}+\sigma_2\mathbbm{I}_{t\geq{}t^*})dW_t
\end{gather}
Up to the time $t^*$, the above model is identical to \eqref{gaussian_sde} and the particle filter performs as demonstrated in the previous chapter. In the case of the Liu and West filter, the range of the posterior support will narrow in line with the converging theoretical posterior as the number of observations increase up to time $t^*$. At the point $t=t^*$, two situations are possible, the new value $\sigma_2$ could lie either inside or outside the range of the estimated posterior. In the case that it is inside, that is there are at least two particles such that $\sigma_t^{(i)}\leq\sigma_2\leq\sigma_t^{(j)}$, the weights of the particles closest to $\sigma_2$ will start to increase and eventually the filter will converge to the new value. However, in order to develop and illustrate a genetic algorithm approach, this section will focus on the opposite case, where $\sigma_2$ is outside the range of the posterior. This situation will be labelled the adaptation phase.
\paragraph{}
In general, the posterior density, given enough observations, tends to converge around the parameter values set in the simulation used to generate the observations. However, this is not possible during the adaptation phase, since by definition the range of the posterior does not encompass the new parameter value. In this case, the posterior will converge to the point closest to the new value $\sigma_2$, located at the boundary of the existing posterior range. In the limit, all density will be focused on the single particle closest to $\sigma_2$, that is:
\begin{gather}
\pi{}^{(i)}_t\xrightarrow{p}1 \; \text{ for } \;i\; s.t. \; |\sigma_2-\sigma^{(i)}_t|=\inf_{1\leq j\leq N}\lbrack{}|\sigma{}_2-\sigma_t^{(j)}|\rbrack{}
\end{gather}
The presence of random perturbation, in the form of the kernel used in the Liu and West filter, allows the posterior interval to expand and therefore decrease the distance of the interval boundary to $\sigma{}_2$. That is, there is a non-zero probability that the random perturbation results in at least one of the new particle locations falling outside the current estimation boundary. This is especially the case during the adaptation phase, where posterior density is accumulated at the boundary. This translates to
\begin{gather}\label{posterior_expansion}
P(\inf_{1\leq j\leq N}\lbrack{}|\sigma_2-\widetilde{\sigma}_t^{(j)}|\rbrack{}<\inf_{1\leq j\leq N}\lbrack{}|\sigma_2-\sigma_t^{(j)}|\rbrack{})>0, \text{ where }\; \widetilde{\sigma}_t^{(j)}\sim{}{\mathcal{N}}(\sigma_t^{(j)}|m_t^{(j)},h^2V_t)
\end{gather}
The random perturbation combines with the re-selection to form a genetic algorithm capable of adapting the posterior to the new value by expanding the posterior such that $\inf\lbrack{}|\sigma{}_2-\sigma_t^{(j)}|\rbrack{}\rightarrow{}0$, thereby allowing the posterior to shift towards $\sigma{}_2$.
In the Liu and West filter, the kernel variance $V_t$ is determined by the variance of the particles, which tends towards zero as the particles become increasingly concentrated around the boundary. The variance is prevented from reaching zero by the random expansion of the boundary, giving the posterior incremental space preventing collapse to a single point. The net result is a situation where the two forces tend to balance out resulting in a relatively steady rate of boundary expansion towards $\sigma_2$.
This is evident in Figure \ref{t_convergence_regime_lw}: after the regime change there is a slow and relatively constant change in the estimate in the direction of $\sigma_2$.
\paragraph{}
The adaptation demonstrated in this section stems from the combination of random perturbation via the kernel and re-selection, creating a type of genetic algorithm. Adaptation is evident, however it is very slow; the Liu and West filter was designed to smooth the posterior without causing overdispersion, and not to rapidly adapt to model parameter regime changes.

\subsection{Controlling the rate of adaptation}
\begin{figure}[t]
\centerline{\includegraphics[scale=0.7]{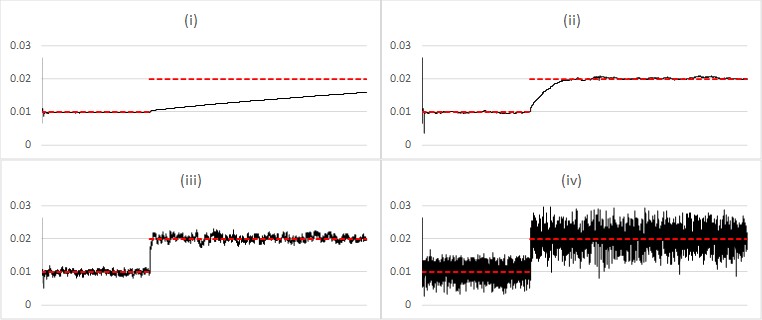}}
\caption{Comparing estimated posterior expected value (black) vs simulation input value (red) of $\sigma_t$, when there is a regime change after 10,000 steps, using the Liu and West filter with additional noise, for different values of $\phi{}$ (i)$\frac{0.1}{N}$ (ii)$\frac{1.0}{N}$ (iii)$\frac{10.0}{N}$ (iv)$\frac{100.0}{N}$}\label{t_convergence_regime_lw_mut}
\end{figure}

\begin{figure}[t]
\centerline{\includegraphics[scale=0.7]{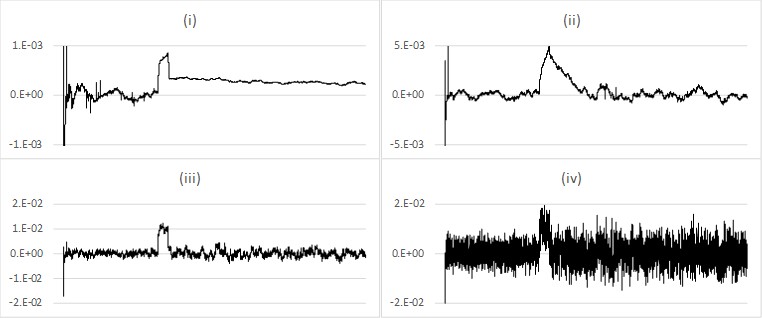}}
\caption{Comparing the change in estimated posterior expected value of $\sigma_t$, where there is a regime change after 10,000 steps, using the Liu and West filter with additional noise, for different values of $\phi{}$ (i)$\frac{0.1}{N}$ (ii)$\frac{1.0}{N}$ (iii)$\frac{10.0}{N}$ (iv)$\frac{100.0}{N}$}\label{t_convergence_regime_lw_mut_diff}
\end{figure}

\begin{figure}[t]
\centerline{\includegraphics[scale=0.7]{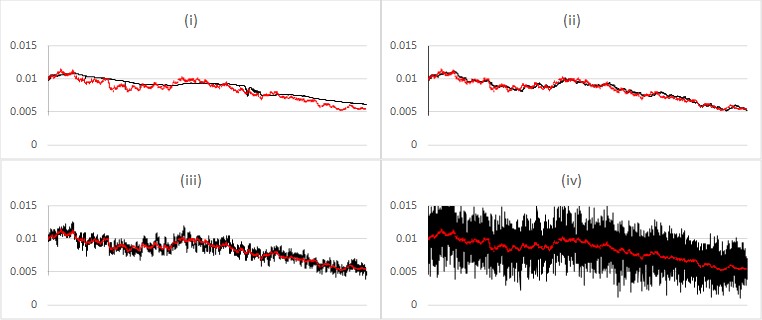}}
\caption{Comparing estimated posterior expected value (black) vs simulation input value (red) of $\alpha$ in the stochastic volatility model, using the Liu and West filter with additional noise, for different values of $\phi{}$ (i)$\frac{0.1}{N}$ (ii)$\frac{1.0}{N}$ (iii)$\frac{10.0}{N}$ (iv)$\frac{100.0}{N}$}\label{t_convergence_stocvol_lw_mut}
\end{figure}

\begin{figure}[t]
\centerline{\includegraphics[scale=0.7]{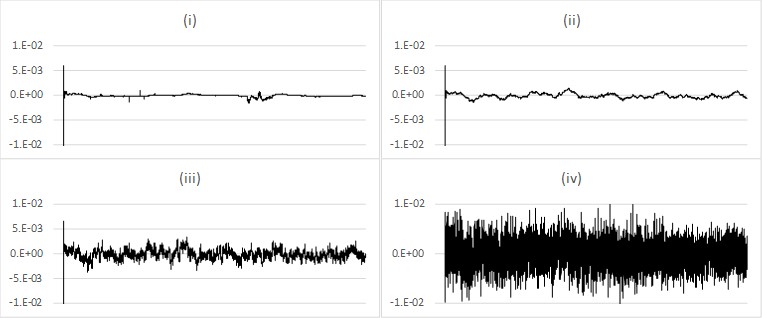}}
\caption{Comparing the change in estimated posterior expected value of $\alpha$ in the stochastic volatility model, using the Liu and West filter with additional noise, for different values of $\phi{}$ (i)$\frac{0.1}{N}$ (ii)$\frac{1.0}{N}$ (iii)$\frac{10.0}{N}$ (iv)$\frac{100.0}{N}$}\label{t_convergence_stocvol_lw_mut_diff}
\end{figure}

Equation \eqref{posterior_expansion} implies that the speed of adaptation is directly related to the speed of expansion of the posterior interval, which in turn is driven by the size of the kernel variance $V_t$. In the case of the Liu and West filter, the adaptation is slow since the variance of the kernel depends on the variance of the posterior and therefore shrinks in line with the convergence of the posterior.
To illustrate the relationship between the size of the random kernel variance and adaptation speed, an additional noise term $\phi$ is introduced into the kernel used by Liu and West as follows:
\begin{gather}\label{LWKernelNoise}
p(\sigma_t|x_{1:t})\approx\sum\limits_{i=1}^N\pi_t^{(i)}{\mathcal{N}}(\sigma_t^{(i)}|m_t^{(i)},h^2V_t+\phi)
\end{gather}
The filtering algorithm becomes:\\[1ex]
\noindent
\fbox{\begin{minipage}{38.5em}
1: \textit{Initialisation} For each particle; let $\sigma_0^{(i)}=\frac{(b-a)i}{N}$ and $\pi_0^{(i)}=\frac{1}{N}$\\
2: Sequentially for each observation:\\
\indent{}	2.1: \textit{Update} For each particle update weight $\hat{\pi}_t^{(i)}=\pi_t^{(i)}p(x_t|x_{t-1},\sigma_t^{(i)})$\\
\indent{}   2.2: \textit{Normalisation} For each particle $\pi_t^{(i)}=\frac{\hat{\pi}_t^{(i)}}{\sum{}\hat{\pi}_t^{(i)}}$\\
\indent{}   2.3: \textit{Resampling} Generate a new set of particles: $$p(\sigma_t|x_{1:t})\approx\sum\limits_{i=1}^N\delta_{\{\sigma_t^{(i)}=\sigma_t\}}\pi_t^{(i)}\xrightarrow[resample]{}p(\sigma_t|x_{1:t})\approx\sum\limits_{k=1}^N\frac{1}{N}\delta_{\{\sigma_t^{(k)}=\sigma_t\}}$$
\indent{}   2.4: \textit{Kernel smoothing} For each particle apply $\sigma_t^{(i)}\sim{}{\mathcal{N}}(\sigma_t^{(i)}|m_t^{(i)},h^2V_t+\phi)$
\end{minipage}}\\[1ex]
\noindent
Figure \ref{t_convergence_regime_lw_mut} and Figure \ref{t_convergence_regime_lw_mut_diff} show filtering results with the above change for varying levels of the noise term $\phi$. The key aspect of the results is that increasing $\phi$ indeed increases the adaptation speed, but at the cost of significant additional noise in the prediction.\footnote{Figure \ref{t_convergence_regime_lw} results are equivalent to the above filter if one sets $\phi=0$.}
\paragraph{}
The regime shift, applied to the dynamics behind Figures \ref{t_convergence_regime_lw_mut} and \ref{t_convergence_regime_lw_mut_diff}, provides a test case with a large sudden change at a specific point in time. As another test case, in Figure \ref{t_convergence_stocvol_lw_mut} and Figure \ref{t_convergence_stocvol_lw_mut_diff} we consider a basic stochastic volatility model. In contrast to a regime shift, changes in the model volatility are driven by a diffusion, testing the capability of the filter to detect continuous, rather than sudden discrete changes. The process is defined by the following system of SDEs:
\begin{gather}\label{stoc_vol_sde}
dx_t=\alpha_tdW_{1,t}\\
d\alpha_t=\nu{}dW_{2,t}
\end{gather}
where $W_{1,t}$ and $W_{2,t}$ denote independent standard Wiener processes.
\paragraph{}
The filter is applied to the stochastic volatility model without any alteration from the setup used to detect regime changes. As is apparent in particular in Figure \ref{t_convergence_stocvol_lw_mut}(ii), some degree of additional noise seems to help in detection of the underlying value of $\alpha$. However, similarly to the regime shift, too much noise simply translates to a noisy estimate.
These results also highlight the resemblance to a filter configured to detect only a stochastic volatility model, i.e., a filter set up to detect the state parameter $\alpha$ given a value of $\nu$\footnote{See \citeasnoun{Bao2012}, \citeasnoun{Casarin2004} for examples of stochastic volatility model detection.}. Each iteration would contain an additional step where each particle's $\alpha_{(i)}$ is updated according to the stochastic volatility dynamic. In this case the additional noise parameter $\phi$ acts in a similar fashion to the stochastic volatility parameter $\nu$. The difference in the approach highlights one of the motivating factors behind our method, the approach being presented does not have to assume prior knowledge of the underlying model, rather it can act as a gauge for empirical assessment of data and with limited modelling assumptions can suggest fruitful extensions toward more sophisticated models.
In the regime shift example the noise parameter $\phi$ improved the adaptation speed at the cost of prediction noise. A high level of $\phi$ is only desirable during the adaptation phase, at other times the ideal the level of $\phi$ would be zero. For the stochastic volatility example clearly there is some optimal level of $\phi$ which achieves good filter performance without causing excessive noise. This is the motivation for a methodology for automatically selecting the level of $\phi$ based on the data, based on an examination of the behaviour of particles on the boundary of the posterior during the adaptation phase, which will be considered next.

\subsection{Applying selection to the rate of adaptation}
\begin{figure}[t]
\centerline{\includegraphics[scale=0.7]{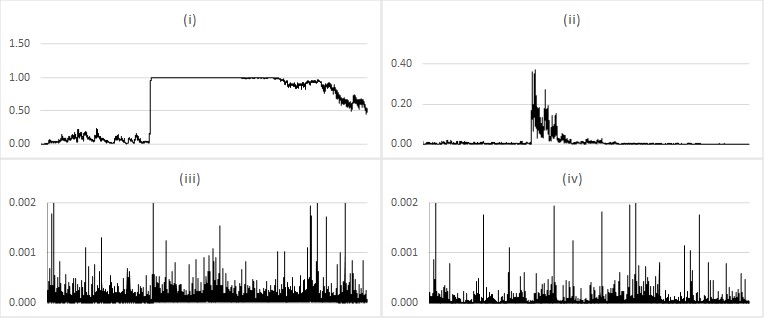}}
\caption{Probability mass shifted beyond the 5\% quantile on the edge of the posterior with regime change after  10,000 steps, using the Liu and West filter with additional noise term $\phi{}$ equal to (i)$\frac{0.1}{N}$ (ii)$\frac{1.0}{N}$ (iii)$\frac{10.0}{N}$ (iv)$\frac{100.0}{N}$}\label{edge_particle_weights}
\end{figure}

\begin{figure}[t]
\centerline{\includegraphics[scale=0.7]{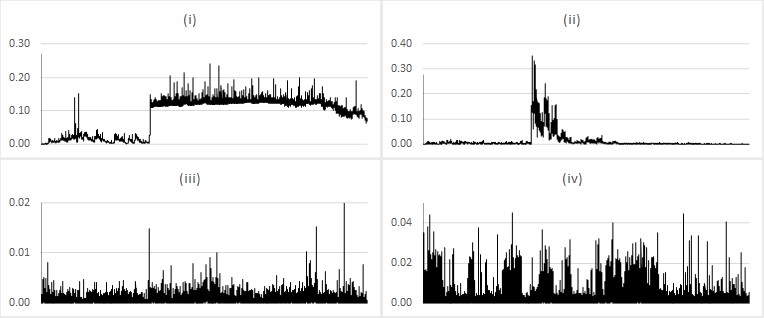}}
\caption{Realised dispersion after re-selection for simulated data with regime change after 10,000 steps, using the Liu and West filter with additional noise term $\phi{}$ equal to (i)$\frac{0.1}{N}$ (ii)$\frac{1.0}{N}$ (iii)$\frac{10.0}{N}$ (iv)$\frac{100.0}{N}$}\label{edge_particle_dispersion_postsel}
\end{figure}

Results from the previous section show that the adaptation of the filter after a regime change is driven by posterior boundary expansion resulting from random perturbation. During the adaptation phase, there is a persistent concentration of density around the boundary of the posterior closest to the new value. It is as though the particles seek to be as close as possible to the new value and are pushing the posterior in this direction. Therefore the behaviour of particles on the edge of the posterior should be quite different during the adaptation phase than at other times. It remains to quantify this difference and use it to enhance the performance of the filter.
\paragraph{}
One of the differences, already highlighted, is the concentration of posterior density around the boundary during the adaptation phase. This is examined numerically by measuring how much probability mass the update step moves into the pre-update tail of the posterior. The measurement is made by first, prior to the update step, finding the lowest $\sigma_t^*$ such that $$\sum\limits_i{}\mathbbm{I}_{(\sigma_t^{(i)}\geq\;\sigma_t^*)}\pi_t^{(i)}\leq{}p$$ when $\sigma_2>\sup\lbrack{}\sigma_t^{(j)}\rbrack{}$ or the highest $\sigma_t^*$ such that $$\sum\limits_i{}\mathbbm{I}_{(\sigma_t^{(i)}\leq\;\sigma_t^*)}\pi_t^{(i)}\leq{}p$$ when $\sigma_2<\inf\lbrack{}\sigma_t^{(j)}\rbrack{}$. Following the update step, compute the amount of probability mass which has moved beyond $\sigma_t^*$, i.e. into the tail, using either $\sum\limits_i{}\mathbbm{I}_{(\sigma_t^{(i)}\geq\;\sigma_t^*)}\hat{\pi}_{t+1}^{(i)}$ when $\sigma_2>\sup\lbrack{}\sigma_t^{(j)}\rbrack{}$ or $\sum\limits_i{}\mathbbm{I}_{(\sigma_t^{(i)}\leq\;\sigma_t^*)}\hat{\pi}_{t+1}^{(i)}$ when $\sigma_2<\inf\lbrack{}\sigma_t^{(j)}\rbrack{}$. If the new cumulative density is higher than $p$, it means that the density in the tail of the posterior has increased. If this measure is persistently high through cycles of weight update and re-selection, it is a strong indicator of regime change.
\paragraph{}
Indeed, Figure \ref{edge_particle_weights} generated with $p=0.05$, reveals a notable increase in the weight associated with the edge particles during the adaption phase. In the first case the measure persists at the maximum value of 1.0, reflecting the slow adaptation observed for this setting, where for a substantial number of update steps all probability mass is shifted beyond $\sigma_t^*$ in each step (i.e., because of the choice of small $\phi$, the posterior moves toward the new ``true value'' only in small increments). Consistent with the findings in the previous sections, the speed of adaptation depends on the size of $\phi$ at the cost of noise in the results.\footnote{Results for Figure \ref{edge_particle_weights} were generated with the identical filter configuration to the results shown in Figure \ref{t_convergence_regime_lw_mut}.}
\paragraph{}
Another quantity to consider is the size of the dispersion of each particle from the application of the kernel. Define realised dispersion for each particle as the distance it has moved from current location due to application of the kernel, denoted as $|\Delta\sigma_t^{(i)}|$.
Consider the situation where all particles are in exactly the same location, i.e the posterior exists at one point. After application of the kernel, it is obvious that the particles on the edge of the posterior will have the highest realised dispersion. In the opposite situation where the particles are very widely dispersed and the kernel variance is relatively small, the relative position of the particle after application of the kernel will have minimal relation with realised dispersion. Therefore the relation between realised dispersion and particle location at the posterior boundary depends on the existing level of dispersion and relative kernel variance. As already determined, during the adaptation phase the particles tend to be very concentrated at the boundary, therefore are closer to the situation where they are likely to exhibit a relation where particles located on the edges will tend to have higher realised dispersion.
\paragraph{}
The combination of higher density and realised dispersion at the edge of the posterior results in a selection bias of high realised dispersion particles during the adaptation phase. This verified numerically by recording the total realised dispersion $\sum\limits_i|\Delta\sigma_t^{(i)}|$ following each re-selection step.
The results are shown in Figure \ref{edge_particle_dispersion_postsel} and show a similar pattern to the results in Figure \ref{edge_particle_weights}, confirming the assertion.

\begin{figure}[t]
\centerline{\includegraphics[scale=0.7]{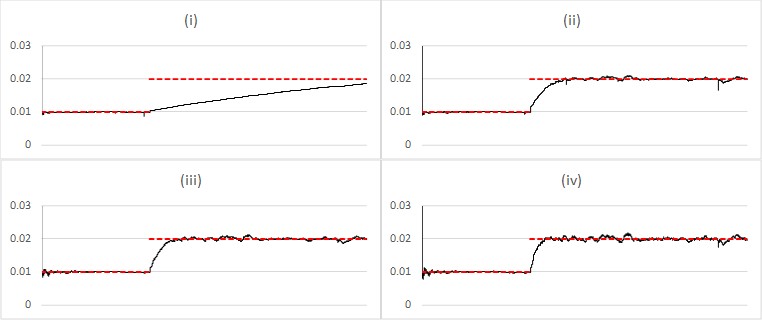}}
\caption{Comparing estimated posterior expected value (black) vs simulation input value (red) of $\sigma_t$ with regime change after 10,000 steps, using the Liu and West filter with additional noise parameter  $\phi^{(i)}\sim{}U(0,c)$ with $c$ set to (i)$\frac{0.2}{N}$ (ii)$\frac{2.0}{N}$ (iii)$\frac{20.0}{N}$ (iv)$\frac{200.0}{N}$}\label{RegimeShift_RandomisedMutation}
\end{figure}

\begin{figure}[t]
\centerline{\includegraphics[scale=0.7]{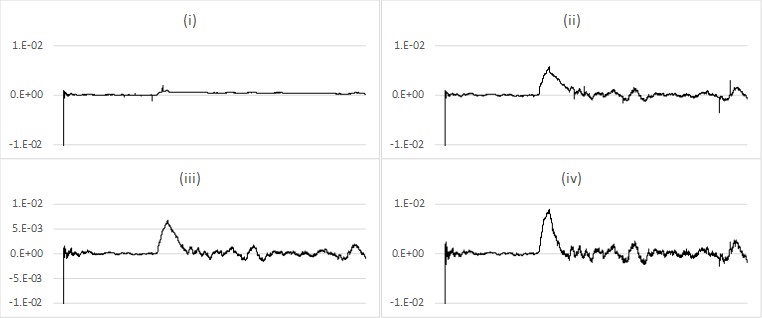}}
\caption{Comparing the change in estimated posterior expected value (black) vs simulation input value (red) of $\sigma_t$ with regime change after 10,000 steps, using the Liu and West filter with additional noise parameter  $\phi^{(i)}\sim{}U(0,c)$ with $c$ set to (i)$\frac{0.2}{N}$ (ii)$\frac{2.0}{N}$ (iii)$\frac{20.0}{N}$ (iv)$\frac{200.0}{N}$}\label{RegimeShift_RandomisedMutation_change}
\end{figure}

\begin{figure}[t]
\centerline{\includegraphics[scale=0.7]{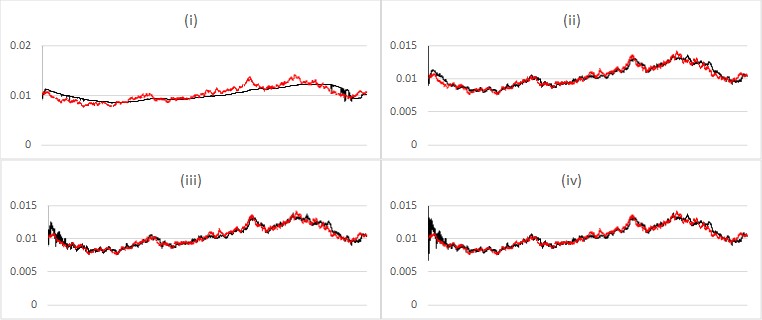}}
\caption{Comparing estimated expected value (black) vs simulation input value (red) of $\alpha$ in the stochastic volatility model for the Liu and West filter with additional noise parameter  $\phi^{(i)}\sim{}U(0,c)$ with $c$ set to (i)$\frac{0.2}{N}$ (ii)$\frac{2.0}{N}$ (iii)$\frac{20.0}{N}$ (iv)$\frac{200.0}{N}$}\label{StocVol_RandomisedMutation}
\end{figure}

\begin{figure}[t]
\centerline{\includegraphics[scale=0.7]{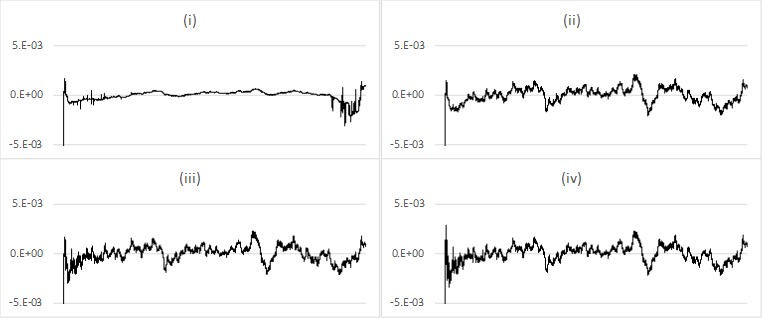}}
\caption{Comparing the change in estimated posterior expected value (black) vs simulation input value (red) of $\alpha$ in the stochastic volatility model, using the Liu and West filter with additional noise parameter  $\phi^{(i)}\sim{}U(0,c)$ with $c$ set to (i)$\frac{0.2}{N}$ (ii)$\frac{2.0}{N}$ (iii)$\frac{20.0}{N}$ (iv)$\frac{200.0}{N}$}\label{StocVol_RandomisedMutation_change}
\end{figure}

The results so far have established a relationship between the value of $\phi$ and the adaptation speed, and a selection bias for particles with high realised dispersion during the adaptation phase. Realised dispersion is a function of $\phi$, which so far has been kept constant, connecting the two results.
Through its connection to realised dispersion, redefining $\phi$ to be non-constant will subject it to the same selection bias. To take advantage of this, define $\phi^{(i)}$ for each particle, initialised using $\phi^{(i)}\sim{}U(0,c)$. This way high values of $\phi$ leading to high dispersion will tend to be selected during the adaptation phase increasing adaptation speed. Conversely, low values of $\phi$ will tend to be selected when adaptation is not required, reducing noise. The filtering algorithm now becomes:\\[1ex]
\noindent
\fbox{\begin{minipage}{38.5em}
1: \textit{Initialisation} For each particle; let $\sigma_0^{(i)}=\frac{(b-a)i}{N}$, $\pi_0^{(i)}=\frac{1}{N}$ and $\phi^{(i)}\sim{}U(0,c)$\\
2: Sequentially for each observation:\\
\indent{}	2.1: \textit{Update} For each particle update weight $\hat{\pi}_t^{(i)}=\pi_t^{(i)}p(x_t|x_{t-1},\sigma_t^{(i)})$\\
\indent{}   2.2: \textit{Normalisation} For each particle $\pi_t^{(i)}=\frac{\hat{\pi}_t^{(i)}}{\sum{}\hat{\pi}_t^{(i)}}$\\
\indent{}   2.3: \textit{Resampling} Generate a new set of particles: $$p(\sigma_t|x_{1:t})\approx\sum\limits_{i=1}^N\delta_{\{\sigma_t^{(i)}=\sigma_t\}}\pi_t^{(i)}\xrightarrow[resample]{}p(\sigma_t|x_{1:t})\approx\sum\limits_{k=1}^N\frac{1}{N}\delta_{\{\sigma_t^{(k)}=\sigma_t\}}$$
\indent{}   2.4: \textit{Kernel smoothing} For each particle apply $\sigma_t^{(i)}\sim{}{\mathcal{N}}(\sigma_t^{(i)}|m_t^{(i)},h^2V_t+\phi^{(i)})$
\end{minipage}}\\[1ex]
The algorithm is tested with the initial distribution set such that the expected value of $\phi$ for each test is equivalent to the value set for the tests in the previous section. The results, shown in Figure \ref{RegimeShift_RandomisedMutation} and Figure \ref{RegimeShift_RandomisedMutation_change}, when compared to Figures \ref{t_convergence_regime_lw_mut} and \ref{t_convergence_regime_lw_mut_diff}, reveal a significant reduction in noise coupled with an increase in the speed for charts (i) and (ii) but a decrease for charts (iii) and (iv). The reduction in noise results from a selection bias for low $\phi$ particles when not in the adaptation phase as discussed above. Conversely the increase in adaptation speed for charts (i) and (ii) results from a selection bias towards higher $\phi$ particles during the adaptation phase. The slow down in adaptation speed observed in charts (iii) and (iv) occurs because prior to the adaptation phase high $\phi$ particles tend to be eliminated from the particle population by the selection process.
The filter is also applied to the stochastic volatility model with results shown in Figure \ref{StocVol_RandomisedMutation} and Figure \ref{StocVol_RandomisedMutation_change}. Similarly to the results for the regime change there is an elimination of noise from the results. However the results, particularly for charts (ii), (iii) and (iv) are very similar to each other indicating that the selection process has converged on a similar level of $\alpha$, highlighting the ability of the filter to find the correct level of additional noise corresponding to the constant stochastic volatility parameter.
\paragraph{}
The above algorithm takes advantage of the existing selection process to increase the adaptation speed when required and reduce noise in the results when adaptation is not required. The detection of and increase in adaptation speed during the adaptation phase is now embedded in the algorithm via the selection of the noise term. However the speed of adaptation remains relatively constant, bounded by the range of the initial distribution of $\phi$, which can only shrink as a result of the selection process. The next section describes a method which overcomes this limitation and achieves accelerated adaptation.
\subsection{Accelerated  adaptation: selectively increasing the rate of adaptation}
\begin{figure}[t]
\centerline{\includegraphics[scale=0.7]{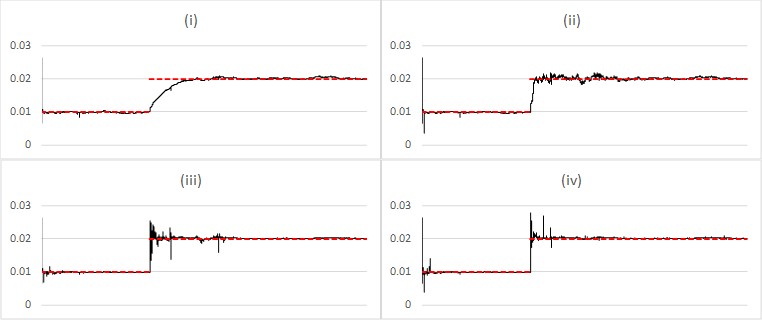}}
\caption{Comparing estimated posterior expected value (black) vs simulation input value (red) of $\sigma$ with regime change after 10,000 steps, using the Liu and West filter with learning for different values of $\gamma$ (i)0.0001 (ii)0.001 (iii)0.01 (iv)0.1}\label{t_convergence_regime_lw_learn}
\end{figure}

\begin{figure}[t]
\centerline{\includegraphics[scale=0.7]{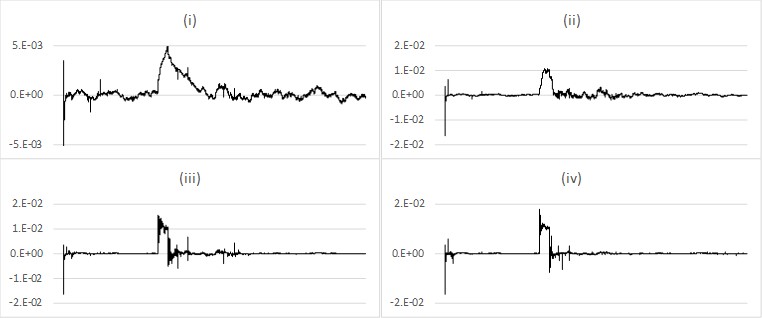}}
\caption{Comparing the change in estimated posterior expected value (black) vs simulation input value (red) of $\sigma$ with regime change after 10,000 steps, using the Liu and West filter with learning for different values of $\gamma$ (i)0.0001 (ii)0.001 (iii)0.01 (iv)0.1}\label{t_convergence_regime_lw_learn_diff}
\end{figure}

\begin{figure}[t]
\centerline{\includegraphics[scale=0.7]{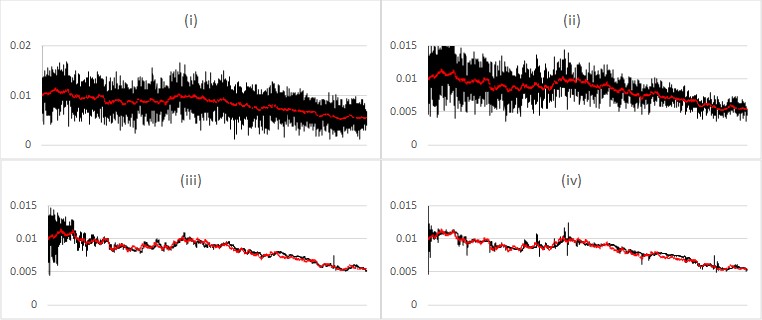}}
\caption{Comparing estimated posterior expected value (black) vs simulation input value (red) of $\sigma$ for stochastic volatility with learning for different values of $\gamma$ (i)0.0001 (ii)0.001 (iii)0.01 (iv)0.1}\label{t_convergence_stocvol_lw_learning}
\end{figure}

\begin{figure}[t]
\centerline{\includegraphics[scale=0.7]{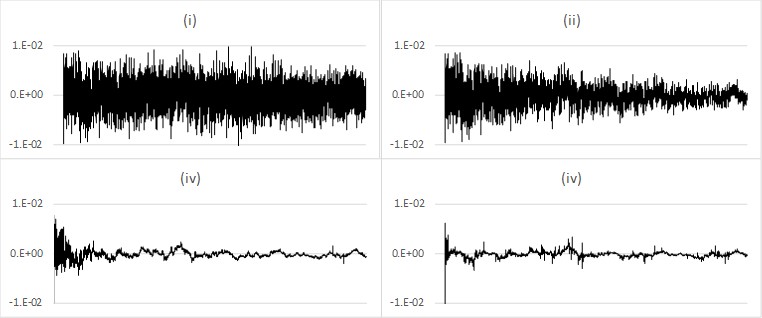}}
\caption{Comparing estimated posterior expected value (black) vs simulation input value (red) of $\sigma$ for stochastic volatility with learning for different values of $\gamma$ (i)0.0001 (ii)0.001 (iii)0.01 (iv)0.1}\label{t_convergence_stocvol_lw_learning_diff}
\end{figure}

Adaptation in a particle filter is driven by a genetic algorithm resulting from a combination of selection and random perturbation. The speed of the adaptation is bound by the size of the parameter $\phi$, which sets level of variance of the random perturbation via the smoothing kernel. To increase the speed of, or accelerate, the rate of adaptation, the parameter $\phi$ needs to constantly increase during the adaptation phase. The idea to allow $\phi$ itself to adapt this way, is to use the already existing genetic algorithm by subjecting $\phi$ to both selection and random perturbation. The effectiveness of selection on $\phi$ has already been demonstrated in the last section. In this section the genetic algorithm for $\phi$ is completed by adding a random perturbation; $\phi_{t+1}^{(i)}=\phi_t^{(i)}e^{\Delta\phi_t^{(i)}}$, where $\Delta\phi_t^{(i)}\sim{}{\mathcal{N}}(0,\gamma)$.
To reflect this, the algorithm is altered as follows:\\[1ex]
\fbox{\begin{minipage}{40.0em}
1: \textit{Initialisation} For each particle; let $\sigma_0^{(i)}=\frac{(b-a)i}{N}$, $\pi_0^{(i)}=\frac{1}{N}$ and $\phi_0^{(i)}\sim{}U(0,c)$\\
2: Sequentially for each observation:\\
\indent{}	2.1: \textit{Update} For each particle update weight $\hat{\pi}_t^{(i)}=\pi_t^{(i)}p(x_t|x_{t-1},\sigma_t^{(i)})$\\
\indent{}   2.2: \textit{Normalisation} For each particle $\pi_t^{(i)}=\frac{\hat{\pi}_t^{(i)}}{\sum{}\hat{\pi}_t^{(i)}}$\\
\indent{}   2.3: \textit{Resampling} Generate a new set of particles: $$p(\sigma_t|x_{1:t})\approx\sum\limits_{i=1}^N\delta_{\{\sigma_t^{(i)}=\sigma_t\}}\pi_t^{(i)}\xrightarrow[resample]{}p(\sigma_t|x_{1:t})\approx\sum\limits_{k=1}^N\frac{1}{N}\delta_{\{\sigma_t^{(k)}=\sigma_t\}}$$
\indent{}   2.4: \textit{Noise parameter perturbation} For each particle; $\phi_t^{(i)}=\phi_{t-1}^{(i)}e^{\Delta\phi_t^{(i)}}$ where $\Delta\phi_t^{(i)}\sim{}{\mathcal{N}}(0,\gamma)$\\
\indent{}   2.5: \textit{Kernel smoothing} For each particle apply $\sigma_t^{(i)}\sim{}{\mathcal{N}}(\sigma_t^{(i)}|m_t^{(i)},h^2V_t+\phi^{(i)})$
\end{minipage}}\\[1ex]
The ability of the proposed approach to accelerate adaptation is demonstrated by adding noise parameter perturbation to the filter configuration used to produce the results shown in Figure \ref{RegimeShift_RandomisedMutation}, chart (ii). The results, shown in Figure \ref{t_convergence_regime_lw_learn} and Figure \ref{t_convergence_regime_lw_learn_diff} for increasing values of $\gamma$, demonstrate very effective acceleration of adaptation coupled with a significant reduction in noise compared to the implementation in the previous section.
The adaptation of the filter to a regime change demonstrates the ability to rapidly increase the speed of adaptation when required. The same mechanism forces the additional noise to decrease when the adaptation phase is finished. The decrease in the noise factor can be further demonstrated with stochastic volatility model simulated data and the filter configured with a very high starting point for $\phi$ as in  Figure \ref{t_convergence_stocvol_lw_mut}, chart (iv). The results in Figure \ref{t_convergence_stocvol_lw_learning} and Figure \ref{t_convergence_stocvol_lw_learning_diff} demonstrate how the particle filter learns to reduce excess noise for increasing values of $\gamma$.
It is also evident from the results, particularly Figure \ref{t_convergence_regime_lw_learn_diff}, that the reduction in noise post adaptation phase tends to be slower than the initial increase. To speed up this reversal a dampening parameter is introduced in the next section.

\subsection{Dampening the rate of adaptation}
\begin{figure}[t]
\centerline{\includegraphics[scale=0.7]{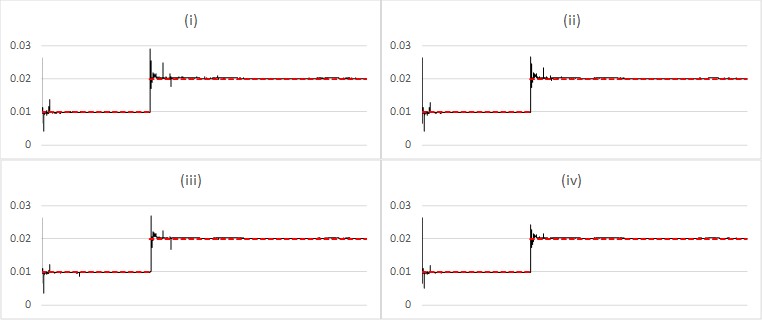}}
\caption{Comparing estimated posterior expected value (black) vs simulation input value (red) of $\sigma$ with regime change after 10,000 steps, using the Liu and West filter with learning for different values of $\kappa$ (i)0.01 (ii)0.02 (iii)0.03 (iv)0.04}\label{t_convergence_regime_lw_learn_damp}
\end{figure}

\begin{figure}[t]
\centerline{\includegraphics[scale=0.7]{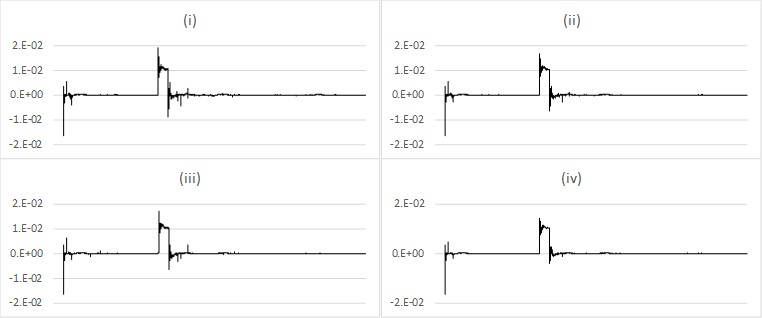}}
\caption{Comparing the change in estimated posterior expected value (black) vs simulation input value (red) of $\sigma$ with regime change after 10,000 steps, using the Liu and West filter with learning for different values of $\kappa$ (i)0.01 (ii)0.02 (iii)0.03 (iv)0.04}\label{t_convergence_regime_lw_learn_damp_diff}
\end{figure}

The post-adaptation learning parameter decrease tends to be slower than the adaptation increase because, relatively speaking, large observed changes are less likely assuming a low volatility than small observed changes assuming a high volatility. Therefore, during adaptation low noise particles are relatively less likely to survive than high noise particles outside of the adaptation phase.
This bias can be counteracted with the introduction of a dampening parameter in the form of a negative mean in the distribution used to perturb $\phi$, that is $\Delta\phi_t^{(i)}$ in the learning step becomes $\Delta\phi_t^{(i)}\sim{}{\mathcal{N}}(-\kappa,\gamma)$.
The addition of the dampening parameter also speeds up the convergence to the Liu and West filter when learning is not required, that is, in the idealised situation where the model assumed by the filter actually matches the observations. The filtering algorithm including the dampening factor becomes:\\[1ex]
\noindent
\fbox{\begin{minipage}{40.0em}
1: \textit{Initialisation} For each particle; let $\sigma_0^{(i)}=\frac{(b-a)i}{N}$, $\pi_0^{(i)}=\frac{1}{N}$ and $\phi_0^{(i)}\sim{}U(0,c)$\\
2: Sequentially for each observation:\\
\indent{}	2.1: \textit{Update} For each particle update weight $\hat{\pi}_t^{(i)}=\pi_t^{(i)}p(x_t|x_{t-1},\sigma_t^{(i)})$\\
\indent{}   2.2: \textit{Normalisation} For each particle $\pi_t^{(i)}=\frac{\hat{\pi}_t^{(i)}}{\sum{}\hat{\pi}_t^{(i)}}$\\
\indent{}   2.3: \textit{Resampling} Generate a new set of particles: $$p(\sigma_t|x_{1:t})\approx\sum\limits_{i=1}^N\delta_{\{\sigma_t^{(i)}=\sigma_t\}}\pi_t^{(i)}\xrightarrow[resample]{}p(\sigma_t|x_{1:t})\approx\sum\limits_{k=1}^N\frac{1}{N}\delta_{\{\sigma_t^{(k)}=\sigma_t\}}$$
\indent{}   2.4: \textit{Noise parameter perturbation} For each particle; $\phi_t^{(i)}=\phi_{t-1}^{(i)}e^{\Delta\phi_t^{(i)}}$ where $\Delta\phi_t^{(i)}\sim{}{\mathcal{N}}(\kappa,\gamma)$\\
\indent{}   2.5: \textit{Kernel smoothing} For each particle apply $\sigma_t^{(i)}\sim{}{\mathcal{N}}(\sigma_t^{(i)}|m_t^{(i)},h^2V_t+\phi^{(i)})$
\end{minipage}}\\[1ex]
The impact of the dampening parameter was tested on the regime shift data with the particle filter configured with a high learning parameter used to generate the results in chart (iv) in Figure \ref{t_convergence_regime_lw_learn}. The results for increasing value of the dampening parameter are shown in Figure \ref{t_convergence_regime_lw_learn_damp} and Figure \ref{t_convergence_regime_lw_learn_damp_diff}. The dampening parameter does indeed reduce estimate noise, however it is important to note that the dampening parameter has to be set low enough as not to completely offset the impact from perturbation. It may also be possible to evolve this parameter in the same manner as the noise parameter. However, this is not attempted in this research.

\subsection{Average perturbation as a relative measure}
The particle filter proposed in this paper allows rapid detection of parameter changes by exploiting and enhancing the genetic algorithm aspect of a filter which includes random perturbation and selection. However, every random perturbation results in a deterioration of the quality of the posterior estimation, since the underlying assumption in the recursive calculation of the particle weights is that the parameters of each particle are fixed. Ideally, if the model assumption in the filter reflected the empirical data, the posterior estimation would not require any additional noise. This leads to the idea that the amount of additional noise used by the filter can serve as an indicator of model adequacy, as well as distinguish between different dynamics present in the data set.
Define this measure as the average of the $\phi$ parameter calculated at each iteration: $$\frac{\sum\limits_i\phi^{(i)}}{N}$$
The following offer some examples of how the behaviour of average $\phi$ can help to distinguish and identify the dynamics of the underlying data.
\subsubsection{Gaussian process}
\begin{figure}[t]
\centerline{\includegraphics[scale=0.7]{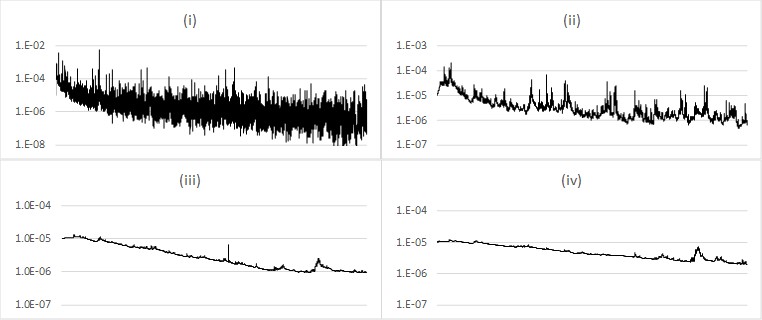}}
\caption{Comparison of average $\phi$ (log-scale) for the Gaussian model with different values of filter perturbation variance $\gamma$  (i) 1.0 (ii) 0.01 (iii) 0.001 (iv) 0.0001}\label{GaussianLearningMeasure}
\end{figure}

The Gaussian process as the underlying dynamic demonstrates the behaviour of the measure when the data generating process matches the assumption in the filter. As the estimated parameter posterior converges increasingly less perturbation is required, reflecting the correspondence between the filter assumption and underlying data. Simulation results shown in Figure \ref{GaussianLearningMeasure} confirm the convergence of $\phi$ for different perturbation variance parameter $\gamma$, expectedly high $\gamma$ results in convergence noise highlighting the need for some implementation specific tuning of this parameter.

\subsubsection{Regime change}
\begin{figure}[t]
\centerline{\includegraphics[scale=0.7]{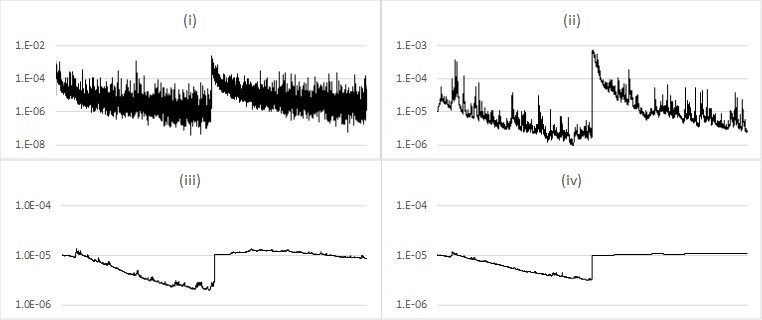}}
\caption{Comparison of average $\phi$ (log-scale) for the regime change model with different values of filter perturbation variance parameter $\gamma$  (i) 1.0 (ii) 0.01 (iii) 0.001 (iv) 0.0001}\label{RegimeLearningMeasure}
\end{figure}

The regime change is marked by a sharp increase in $\phi$, reflecting sudden adaptation to the new model state. Before and after the regime change the model is Gaussian and therefore the behaviour of $\phi$ is similar to the previous section. The rate of convergence of $\phi$ is slower for the higher $\sigma$ following the regime change indicating a relationship between the rate of convergence of $\phi$ and $\sigma$. The results are shown in Figure \ref{RegimeLearningMeasure} for varying levels of $\gamma$.

\subsubsection{Stochastic volatility}
\begin{figure}[t]
\centerline{\includegraphics[scale=0.7]{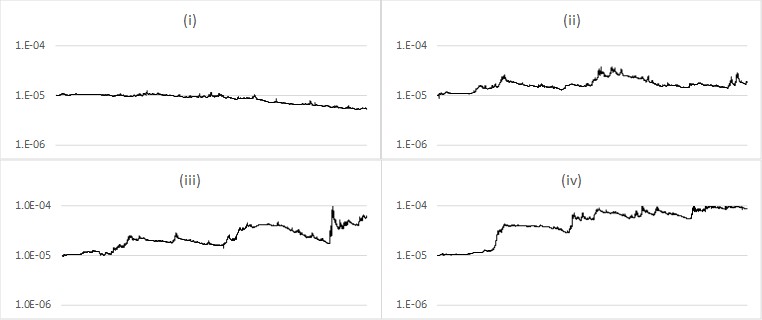}}
\caption{Comparison of average $\phi$ (log-scale) for the stochastic volatility model with different values of stochastic volatility $\nu$  (i) 0.1 (ii) 0.2 (iii) 0.3 (iv) 0.4}\label{StocVolLearningMeasure}
\end{figure}

If the filter assumption does not match the dynamics of the underlying data, $\phi$ will not tend to converge to zero. In the case of stochastic volatility, $\phi$ will tend towards a constant value reflecting the constantly changing volatility, with the level of this value indicating the level of stochasticity in the data. Figure \ref{StocVolLearningMeasure} shows some examples of the behaviour of $\phi$ for varying levels of stochasticity in the data, using $\gamma=0.001$ corresponding to chart (iii) in Figure \ref{GaussianLearningMeasure}.

\section{Conclusion}
\paragraph{}
We arrived at our proposed methodology by first recognising that the random perturbation technique applied in a particle filter results in a genetic--type algorithm capable of adapting to changing parameters. At this point, we took an opposite direction to the approach of \cite{Liu2001}, instead of remediating the overdispersion caused by random perturbation, we allowed the random perturbation to freely evolve, enhancing the adaptive capability of the particle filter. Our approach is highly adaptive when required and convergent conditional on the data matching modelling assumptions and no parameter changes. Given that the level of adaptability is governed by the variance of the random perturbation; the key insight of our approach is that an effective way of recognising the level of required variance is to incorporate its selection into the already existing genetic algorithm framework. In terms of existing literature, it links particle filtering with genetic algorithms for parameter learning, resulting in a filtering algorithm particularly useful for parameter change detection and in the context of finance an effective on-line method for measuring volatility.

\paragraph{}
As with particle filters in general our approach, is easy to implement requiring a relatively small amount of coding and our results therefore are easy to replicate. We also demonstrate our results with a basic model adding to the ease of replication of the methodology. We provided an implementation orientated overview by presenting our approach in an incremental fashion, following chronological developments in particle filtering leading up to our contribution. In this way we hope to accommodate an audience interested in our approach with various degrees of experience in particle filtering, particularly in finance research and practice where particle filtering is not widely used.

\bibliography{../BibMaster}
\end{document}